\documentclass{article}

\PassOptionsToPackage{numbers, compress}{natbib}
\usepackage[preprint]{neurips_2024}

\usepackage[utf8]{inputenc} %
\usepackage[T1]{fontenc}    %
\usepackage{hyperref}       %
\usepackage{url}            %
\usepackage{booktabs}       %
\usepackage{amsfonts}       %
\usepackage{nicefrac}       %
\usepackage{microtype}      %
\usepackage{pifont}%
\usepackage{graphicx}
\usepackage{amsmath, amsthm, amssymb, bm}
\usepackage{color, colortbl}
\usepackage[dvipsnames]{xcolor}
\usepackage{mathtools}
\usepackage{enumitem}
\usepackage{multirow}
\usepackage[ruled,vlined]{algorithm2e}
\usepackage{caption}
\usepackage{subcaption}
\usepackage{booktabs}
\usepackage{cleveref}
\usepackage{listings}
\usepackage{xcolor}
\usepackage{graphicx,wrapfig,lipsum}
\usepackage{wrapfig}
\usepackage{tabularx}

\usepackage{graphicx}
\usepackage{booktabs}
\usepackage{abbrv}
\usepackage{tcolorbox}
\usepackage{varwidth}

\newcommand{\VarSty}[1]{\begin{varwidth}{\linewidth}#1\end{varwidth}}

\usepackage[accsupp]{axessibility}  %

\usepackage{orcidlink}
\usepackage{threeparttable}
\usepackage{listings}
\usepackage{xcolor}
\usepackage{amsmath} %

\newcommand{\algo}{\texttt{Piva}\xspace}
\newcommand{\longalgo}{\textbf{P}reserving \textbf{I}dentity with \textbf{Va}riational Score Distillation}
\newcommand{\benchmark}{\texttt{Noe}\xspace}
\newcommand{\longbenchmark}{\textbf{N}erf-synthetic and \textbf{O}bjaverse synthetic 3D \textbf{E}diting Benchmark}
\usepackage[colorinlistoftodos, textwidth=30mm, shadow, textsize=small]{todonotes}
\definecolor{babyblue}{rgb}{0.54, 0.81, 0.94}
\definecolor{citrine}{rgb}{0.89, 0.82, 0.04}
\definecolor{misocolor}{rgb}{0.16,0.27,0.86}
\definecolor{jbcolor}{rgb}{0.9,0.4,0.2}
\definecolor{bernacolor}{rgb}{0.9608,0.4863,0.00}
\definecolor{carlcolor}{rgb}{0.0,0.9863,0.30}
\definecolor{grey}{rgb}{0.3, 0.3, 0.3}

\definecolor{graphicbackground}{rgb}{0.96,0.96,0.8}
\definecolor{rouge1}{RGB}{226,0,38}  %
\definecolor{orange1}{RGB}{243,154,38}  %
\definecolor{jaune}{RGB}{254,205,27}  %
\definecolor{blanc}{RGB}{255,255,255} %
\definecolor{rouge2}{RGB}{230,68,57}  %
\definecolor{orange2}{RGB}{236,117,40}  %
\definecolor{taupe}{RGB}{134,113,127} %
\definecolor{gris}{RGB}{91,94,111} %
\definecolor{bleu1}{RGB}{38,109,131} %
\definecolor{bleu2}{RGB}{28,50,114} %
\definecolor{vert1}{RGB}{133,146,66} %
\definecolor{vert3}{RGB}{20,200,66} %
\definecolor{vert2}{RGB}{157,193,7} %
\definecolor{darkyellow}{RGB}{233,165,0}  %
\definecolor{lightgray}{rgb}{0.9,0.9,0.9}
\definecolor{darkgray}{rgb}{0.6,0.6,0.6}
\definecolor{babyblue}{rgb}{0.54, 0.81, 0.94}
\definecolor{citrine}{rgb}{0.89, 0.82, 0.04}
\definecolor{misogreen}{rgb}{0.25,0.6,0.0}
\definecolor{PalePurp}{rgb}{0.66,0.57,0.66}
\definecolor{todocolor}{rgb}{0.66,0.99,0.99}
\definecolor{pearOne}{HTML}{2C3E50}
\definecolor{pearTwo}{HTML}{A9CF54}
\definecolor{pearTwoT}{HTML}{C2895B}
\definecolor{pearThree}{HTML}{E74C3C}
\colorlet{titleTh}{pearOne}
\colorlet{bull}{pearTwo}
\definecolor{pearcomp}{HTML}{B97E29}
\definecolor{pearFour}{HTML}{588F27}
\definecolor{pearFith}{HTML}{ECF0F1}
\definecolor{pearDark}{HTML}{2980B9}
\definecolor{pearDarker}{HTML}{1D2DEC}
\definecolor{darkTurquoise}{HTML}{007C7D}
\hypersetup{
	colorlinks,
	citecolor=darkTurquoise,
	linkcolor=pearThree,
	breaklinks=true,
	urlcolor=pearDarker}
	
\newcommand{\best}[1]{\textbf{\textcolor{darkTurquoise}{{#1}}}}
\newcommand{\secondbest}[1]{\textcolor{darkTurquoise}{{#1}}}

\title{Preserving Identity with Variational Score for General-purpose 3D Editing}

\author{%
  Duong H. Le$^1$*\quad Tuan Pham$^2$*\quad Aniruddha Kembhavi$^1$\quad Stephan Mandt$^2$\quad \\\textbf{Wei-Chiu Ma}$^{1,3}$\quad \textbf{Jiasen Lu}$^1$\\
  $^1$AI2 \quad $^2$University of California, Irvine \quad $^3$Cornell University \quad *equal contribution
}

\begin{document}

\definecolor{mydarkblue}{rgb}{0,0.08,1}
\definecolor{mydarkgreen}{rgb}{0.02,0.6,0.02}
\definecolor{mydarkred}{rgb}{0.8,0.02,0.02}
\definecolor{mydarkorange}{rgb}{0.40,0.2,0.02}
\definecolor{mypurple}{RGB}{111,0,255}
\definecolor{myred}{rgb}{1.0,0.0,0.0}
\definecolor{mygold}{rgb}{0.75,0.6,0.12}
\definecolor{myblue}{rgb}{0,0.2,0.8}
\definecolor{mydarkgray}{rgb}{0.66,0.66,0.66}

\newcommand{\duong}[1]{\textcolor{mydarkblue}{[Duong: #1]}}
\newcommand{\weichiu}[1]{\textcolor{red}{Wei-Chiu: #1}}
\newcommand{\jiasen}[1]{\textcolor{cyan}{Jiasen: #1}}

\colorlet{punct}{red!60!black}
\definecolor{background}{HTML}{EEEEEE}
\definecolor{delim}{RGB}{20,105,176}
\colorlet{numb}{magenta!60!black}

\maketitle

\begin{abstract}
We present \algo (\longalgo), a novel optimization-based method for editing images and 3D models based on diffusion models. 
Specifically, our approach is inspired by the recently proposed method for 2D image editing - Delta Denoising Score (DDS). 
We pinpoint the limitations in DDS for 2D and 3D editing, which causes detail loss and over-saturation. 
To address this, we propose an additional score distillation term that enforces identity preservation. This results in a more stable editing process, gradually optimizing NeRF models to match target prompts while retaining crucial input characteristics.
We demonstrate the effectiveness of our approach in zero-shot image and neural field editing. 
Our method successfully alters visual attributes, adds both subtle and substantial structural elements, translates shapes, and achieves competitive results on standard 2D and 3D editing benchmarks.
Additionally, our method imposes no constraints like masking or pre-training, making it compatible with a wide range of pre-trained diffusion models. This allows for versatile editing without needing neural field-to-mesh conversion, offering a more user-friendly experience.
\end{abstract}

\section{Introduction}

The importance of 3D editing is growing across various fields, from virtual reality and gaming to medical imaging and architectural visualization. Manipulating three-dimensional objects precisely and flexibly is critical to creating engaging and realistic digital experiences. Traditionally, 3D editing has depended on manual techniques and specialized software, which are often time-consuming and require significant expertise.

Recently, alternative methods for 3D editing have emerged due to breakthroughs in vision and language foundation models like Stable Diffusion \cite{rombach2022high} and DALL-E 3, which allow for editing both 2D and 3D assets using text prompts. One such method is the Delta Denoising Score (DDS) \cite{hertz2023delta}, which enables zero-shot editing through parametric optimization. However, DDS is unstable and can significantly alter the edited image from the original, sometimes resulting in overly saturated results.

With these challenges in mind, we introduce \algo, \longalgo, a novel and effective approach to textual neural editing in 3D. \algo maintains the identity of the source by integrating an additional score distillation term to the DDS loss. This term minimizes the differences between distributions of rendered images from the original and the edited NeRF, thus preserving the source identities.

Combining this term with the DDS allows us to edit and produce high-quality 3D scenes/assets from textual descriptions. We demonstrate the effectiveness of our approach in both zero-shot 2D and 3D editing. Our approach effectively edits high-quality synthetic objects from Objaverse \cite{deitke2023objaverse} and real-world scenes. Precisely, we can edit the geometry of a given model or add a new object to the scene with negligible changes in irrelevant parts.

In addition, we observe the lack of a standardized benchmark for general-purpose, non-mask 3D editing. To address that, we introduce \benchmark (\longbenchmark) 
with $35$ objects spanning across multiple categories. For each object, we target $3$ types of editing: changing the appearance with negligible changes in geometry, adding part/object, and (global) translation. In total, the \benchmark benchmark comprises $112$ editing entries, each consisting of the source prompt, target prompt, and instruction. In summary, we make the following contributions:
\begin{enumerate}
    \item We propose \algo, a method that improves upon DDS. \algo allows general-purpose 3D editing with user-friendly interface \ie only target and source prompt are required.
    \item We introduce a novel benchmark for text-based 3D editing on synthetic objects with diverse and complex editing scenarios.
    \item Through comprehensive experiments, we empirically show that \algo can produce high-quality editing that aligns with the target text prompt while maintaining the key characteristic of the source input.
\end{enumerate}

\section{Preliminary}
\label{sec:background}
We present preliminaries on diffusion models, score distillation sampling, variational score distillation, and delta-denosing score.

\noindent \textbf{Diffusion models.} 
A diffusion model \cite{sohl2015deep, ho2020denoising, song2020score} consists of a forward process $q$ that gradually adds noise into a data point $\mathrm{x}$, and a reverse process $p$ that slowly denoise from the random noise $\mathbf{z}_t$. The forward process is defined by $q(\mathbf{x}_t | \mathbf{x}) = \mathcal{N}\left(\alpha_t \mathbf{x}, \sigma_t^2 \mathbf{I}\right)$, in which $\alpha_t, \sigma_t > 0$ are hyperparameters satisfying $\alpha_0 \approx	
1, \sigma_0 \approx	0, \alpha_T \approx 0, \sigma_T \approx 1$
; and the reverse process is defined by denoising from 
$p(\mathbf{x}_T) = \mathcal{N}(0, \mathbf{I})$ to the data distribution.
The denoising network $\epsilon_\kappa\left(\mathbf{x}_t ; t\right)$ is trained by minimizing
\begin{equation}
\label{eq:ddpm_loss}
\mathcal{L}_{\texttt{diff }}(\kappa, \mathbf{x})=\mathbb{E}_{t, \epsilon}\left[\omega(t)\left\|\epsilon_\kappa\left(\alpha_t \mathbf{x}+\sigma_t \epsilon ; t\right)-\epsilon\right\|_2^2\right]
\end{equation}
where $\omega(t)$ is a time-dependent weighting function. After training, we have $p_t \approx q_t$, and thus we can draw samples from $p_0 \approx q_0$.

\noindent \textbf{Score Distillation Sampling (SDS).} In contrast to the standard diffusion models where sampling over pixel (or latent) space, DreamFusion \cite{poole2022dreamfusion} proposes to sample on the parameter space through their innovative \textit{score distillation sampling} objective. This objective allows optimizing a NeRF (or any differentiable image generator) by differentiating the training objective of the diffusion model with respective to rendered images $\mathbf{x} = g(\theta, \pi)$ where $g$ is the NeRF model and $\pi$ is a sampled camera pose. Particularly, the gradient of the loss function is
\begin{equation}
\label{eq:sds}
\nabla_\theta \mathcal{L}_{\texttt{SDS}}(\theta) = \mathbb{E}_{t,\pi,\epsilon} \left[\omega(t)(\epsilon_\kappa(\mathbf{x}_t, t, \pi, \mathbf{c}) - \epsilon)\frac{\partial g(\theta)}{\partial\theta}\right]
\end{equation}
where $\mathbf{x}_t = \alpha_t g(\theta, \pi) + \sigma_t \mathbf{\epsilon}$ is the noisy rendered image. Note that SDS ignores the U-Net Jacobian term in the gradient calculation. Moreover, they require an unusually high classifier-free guidance (CFG) value to converge. Empirical observations \cite{poole2022dreamfusion} show that SDS often suffers from mode collapse and over-saturation problems.

\noindent \textbf{Variational Score Distillation (VSD).} To address the over-saturation and mode-collapse problem, Wang \etal \cite{wang2023prolificdreamer} propose viewing the distillation process through the lens of a particle-based variational inference framework. Instead of using the injected noise in the gradient as in Equation \ref{eq:sds}, they train a second diffusion model, which is parameterized by $\phi$, to approximate the score of distribution of rendered images of generated NeRF. Formally, the gradient to update NeRF is
\begin{equation}
\label{eq:vsd}
\nabla_\theta \mathcal{L}_{\texttt{VSD}}(\theta, \phi) = \mathbb{E}_{t, \pi,\epsilon} \left[\omega(t)(\epsilon_{\kappa}(\mathbf{x}_t, t, \pi, \mathbf{c}) - \epsilon_\phi(\mathbf{x}_t, t, \pi, \mathbf{c}))\frac{\partial g(\theta)}{\partial\theta}\right]
\end{equation}
While VSD works well with various CFG and improves the diversity and sample quality, it is designed to generate 3D from text. 

\noindent \textbf{Delta Denoising Score (DDS).} Inspired by Dreamfusion \cite{poole2022dreamfusion}, Hertz \etal \cite{hertz2023delta} cast the image editing problem as an optimization where they distill from a pre-trained diffusion model to steer an image towards a desired direction by a text. 
DDS proposes subtracting estimated scores from the reference image to solve the non-detailed and blurry outputs due to noisy gradients produced by SDS.
The editing operation is mask-free, which can bring many advantages, especially for 3D editing, as we shall show. 
The gradient for DDS is
\begin{equation}
\label{eq:dds}
\nabla_\theta \mathcal{L}_{\texttt{DDS}}(\theta) = \mathbb{E}_{t, \pi, \epsilon}\left[\omega(t)(\epsilon_\kappa(\mathbf{x}_t, t, \pi, \mathbf{c}_{tgt}) - \epsilon_\kappa(\mathbf{\hat{x}}_t, t, \pi, \mathbf{c}_{src}))\frac{\partial g(\theta)}{\partial\theta}\right]
\end{equation}
in which $\mathbf{x}_t$ is the noisy edited image and ${\mathbf{\hat{x}}_t}$ is the noisy original image. While DDS can mitigate the inherent blurry effect of SDS, it can not explicitly enforce identity preservation from source images. 
\section{Methodology}
\subsection{Problem formulation and notation}
We are given a differentiable image generator $g(\theta)$ with parameter $\theta$. Specifically, in 3D editing, $g(\theta)$ refers to the NeRF model; and for the 2D case, $g(.)$ is an identity mapping and parameter $\theta$ is the image $\mathbf{x}$. 
Our goal is to edit the NeRF model or image, where we have the target condition \eg text prompt $\mathbf{c}_{tgt}$ to describe the desired results and the source prompt $\mathbf{c}_{src}$ to outline the original ones. Furthermore, we do not assume to access any masks or bounding boxes to specify the editable regions. We denote $\mathbf{x}$ as the edited image and $\mathbf{\hat{x}}$ as the original image. $\mathbf{x}_t$ and $\mathbf{\hat{x}}_t$ are the corresponding noisy version of $\mathbf{x}$ and $\mathbf{\hat{x}}$ at time step $t$.

\subsection{Proposed method}
\label{sec:proposed_method}

\begin{figure}[t]
\centering
\includegraphics[width=\textwidth]{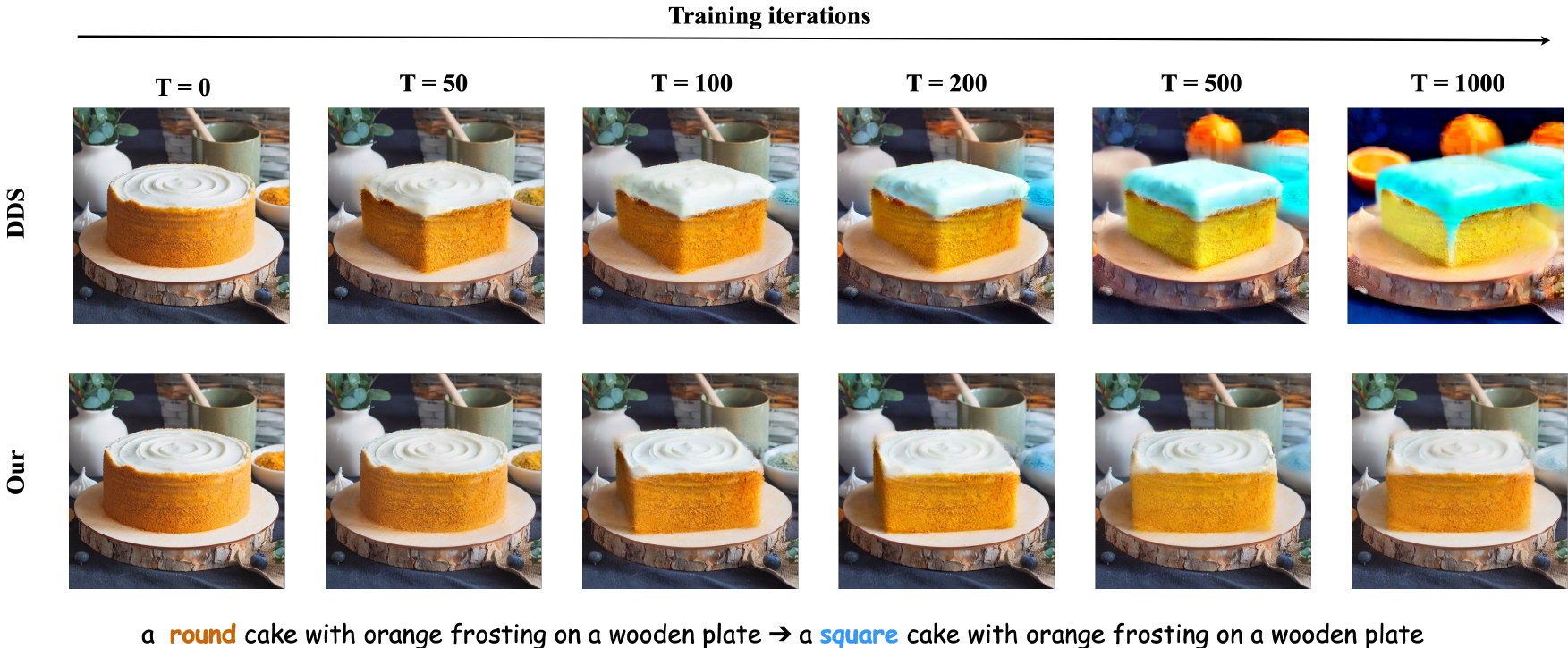}
\caption{Illustrating the content drifting problem of DDS \cite{hertz2023delta} and our method on image editing. While DDS can avoid blurry results and maintain background at the first few training steps, it deviates the input significantly after a few hundred steps and returns over-saturated and irrelevant results. }
\vspace{-3mm}
\label{fig:dds_unstable}
\end{figure}

Here we introduce \longalgo, a method that aims to directly optimize the parametric models (e.g. NeRF) to align with the target condition while keeping other delicate features from the original models intact. 

Our work is inspired by DDS \cite{hertz2023delta}, which helps reduce the noise in the estimated score of pre-trained diffusion models. Consequently, for the editing task, DDS can mitigate the inherent blurry effect of SDS and better preserve the original features, such as background details. In 3D editing, a parametric optimization approach such as DDS has numerous advantages over iterative dataset update methods such as Instruct-NeRF2NeRF \cite{haque2023instruct}. More concretely, direct optimizing the underlying 3D representation enforces consistency across views. On the other hand, Instruct-NeRF2NeRF-like approaches do not guarantee that multiple views share a consistent editing pattern \eg adding a new object to a scene can be tricky as each independent 2D editing operation might result in different positions of added object, thus, the updated dataset comprises inconsistent views. 

Unfortunately, in our experiments, DDS leads to over-saturated images and still deviates significantly from the input. More precisely, while DDS can alleviate the noise in the gradient for updating images, they are imperfect, and errors can accumulate from multiple training iterations, DDS still lacks the ``\emph{reconstruction}'' term for explicitly enforcing identity preservation. Figure \ref{fig:dds_unstable} illustrates the instability of DDS for image editing. Particularly, the optimized image can diverge from the input after a few hundred steps. Although the authors of DDS propose tricks such as early stopping, this instability may hinder the application of DDS in 3D editing. In our experiment, we find that DDS leads to over-saturated results and blurry texture before converging into meaningful/aligned shapes, as in Figure \ref{fig:dds_vs_our}, \ref{fig:dds_vs_our_2d_p2}, \ref{fig:dds_vs_our_3d}.

\begin{figure}[t]
\includegraphics[width=\textwidth]{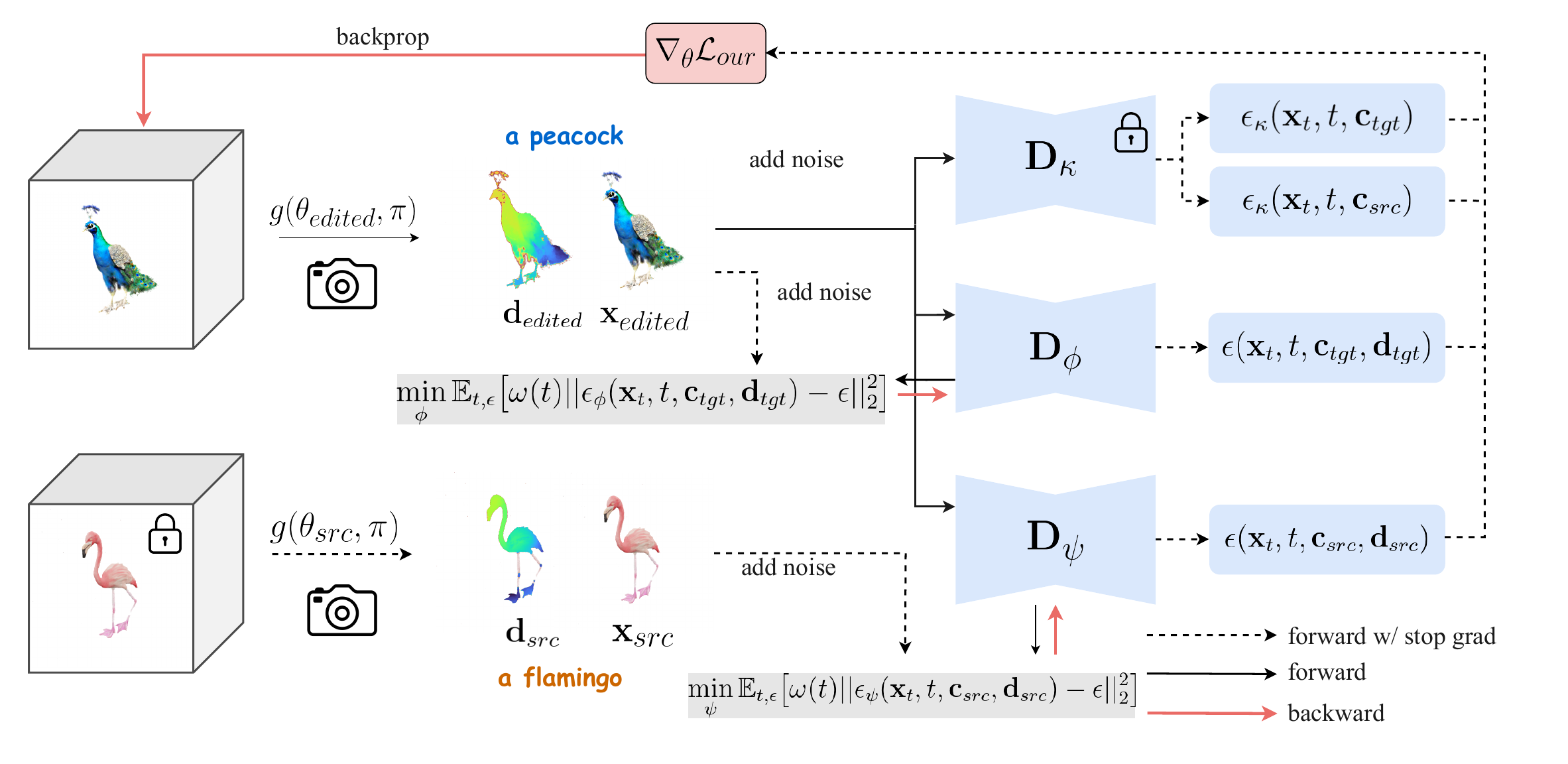}
\vspace{-6mm}
\caption{Illustrating the editing pipeline of our approach. Given a NeRF model, a {target prompt} $\mathbf{c}_{tgt}$ describes the desired editing results and {source prompt} $\mathbf{c}_{src}$ that specify the original NeRF model. We initialize the edited model $\theta_{edited}$ from the input NeRF and update it according to Equation \ref{eq:dds_7}. At the same time, we fine-tune the source and target diffusion model $\mathbf{D}_\phi$ and $\mathbf{D}_\psi$ based on diffusion loss. Thus, the two models can approximate the score of distribution of sampled images from original/edited NeRF.} 
\label{fig:pipeline}
\vspace{-3mm}
\end{figure}

To tackle this problem, we introduce an auxiliary loss that regularizes the output of the DDS during the optimization process. Particularly, we draw inspiration from Diff-Instruct \cite{luo2024diff} and ProlificDreamer \cite{wang2023prolificdreamer} to minimize the KL-divergence between the distribution $p_0$ of the rendered images from the edited NeRF and the distribution $\hat{p}_0$ of the rendered images from the \textbf{original} NeRF. Previous works \cite{luo2024diff, wang2023prolificdreamer} have shown that, to minimize this KL divergence term, we can optimize the model using the following gradient:
\begin{align}
    \nabla_\theta \mathcal{L}(\theta) &= \mathbb{E}_{t, \epsilon} \left[ \omega(t) (\nabla p_t(\mathbf{x}) - \nabla \hat{p}_t(\mathbf{x})) \frac{\partial g(\theta)}{\partial \theta}\right] \label{eq:ikl_grad}
\end{align}
Since the density of $p$ and $\hat{p}$ are intractable, we can utilize a powerful pre-trained text-to-image (T2I) diffusion model to approximate the marginal score of these two distributions. We use a \emph{source diffusion model} $\mathbf{D}_\psi$ to estimate the score $\nabla \hat{p}_t(\mathbf{x})$ of rendered images from \textbf{original} NeRF models. In addition, we use \emph{target diffusion model} $\mathbf{D}_\phi$ to estimate the score $\nabla p_t(\mathbf{x})$ of samples from edited NeRF models. In practice, we use ControlNet \cite{zhang2023adding} with depth condition $\mathbf{d}$ to parameterize $\mathbf{D}_\psi$ and $\mathbf{D}_\phi$; and we apply LoRA finetuning \cite{hu2021lora} for both networks. Equation \ref{eq:ikl_grad} can be rewritten as:
\begin{align}
    \nabla_\theta \mathcal{L}(\theta) &= \mathbb{E}_{t, \epsilon}\left[\omega(t)(\epsilon_\psi(\mathbf{x}_t, t, \mathbf{c}_{src}) - \epsilon_\phi(\mathbf{x}_t, t, \mathbf{c}_{tgt}))\frac{\partial g(\theta)}{\partial\theta} \right] \label{eq:vsd_2}
\end{align}
in which we omitted the camera pose $\pi$. We leverage the conventional diffusion loss for training the target and source diffusion model, \ie, $\mathbf{D}_\phi$ and $\mathbf{D}_\psi$ are optimized to minimize the below objectives:
\vspace{-4mm}

\begin{align}
\label{eq:ddpm_loss}
\mathcal{L}(\psi) &=\mathbb{E}_{t, \epsilon}\left[\omega(t)\left\|\epsilon_\psi\left(\mathbf{\hat{x}}_t ; t\right)-\epsilon\right\|_2^2\right] \\
\label{eq:ddpm_loss_2}
\mathcal{L}(\phi) &=\mathbb{E}_{t, \epsilon}\left[\omega(t)\left\|\epsilon_\phi\left(\mathbf{x}_t ; t\right)-\epsilon\right\|_2^2\right]
\end{align}
An astute reader might find that Equation \ref{eq:vsd_2} resembles the variational score distillation loss \cite{wang2023prolificdreamer}. However, we note that while the VSD tries to distill from the pretrained diffusion model to the rendered image distribution of a NeRF model, we use this score distillation term as a regularizer to help the optimization process of the DDS loss maintain the key characteristic of the original data. Intuitively, the DDS governs the update at the desirable regime while the regularizer helps ``recover'' the edited results if DDS mistakenly deviates edited samples on unwanted regions.

The objective function for editing is a combination of DDS and our auxiliary loss as below:
\begin{align}
\nabla_\theta\mathcal{L}(\theta)
 &= \mathbb{E}_{t, \epsilon}\left[\omega(t)(\underbrace{\epsilon_\kappa(\mathbf{x}_t, t, \mathbf{c}_{tgt}) - \epsilon_\kappa(\mathbf{\hat{x}}_t, t, \mathbf{c}_{src})}_{\texttt{DDS}} \right. \label{eq:dds_7} + \left. \lambda (\underbrace{\epsilon_\psi(\mathbf{x}_t, t, \mathbf{c}_{src}) - \epsilon_\phi(\mathbf{x}_t, t, \mathbf{c}_{tgt}))}_{\texttt{regularizer}}) \frac{\partial g(\theta)}{\partial\theta}\right]
 \end{align}
where $\lambda$ is a hyper-parameter that controls the trade-off between identity preservation and target condition alignment. Perhaps surprisingly, we empirically found that plugging the same noisy latent $\mathbf{x}_t$ in the place of  $\mathbf{\hat{x}_t}$ in Equation \ref{eq:dds_7} could achieve satisfactory results. This reduces our need to render the corresponding views at each optimization step. Thus, we opt-in to use this for all our experiments. The pipeline for our approach is depicted in Figure \ref{fig:pipeline} and outlined in Algorithm \ref{algo}.
\vspace{-2mm}
\subsection{Discussion}
Compared to DDS \cite{hertz2023delta} and VSD \cite{wang2023prolificdreamer}, our method can effectively preserve the identity of the original inputs while aligning it to best match the description as in Figure \ref{fig:dds_vs_our_2d}. Opposite to many prior works in 3D editing that involve a masking procedure \cite{li2023focaldreamer, zhuang2023dreameditor}, we do not require a mask for editing. Thus, our method allows more general types of editing such as global transform \eg a cat to a dog.

In contrast to existing works that leverage the 2D editing diffusion model, such as Instruct-Pix2Pix as in \cite{haque2023instruct},  requires pretraining the editing models, which is time and computationally intensive in situations where we want to adopt a newly released model. Our method does not impose such constraints, thus allowing us to leverage any pretrained model. %
\vspace{-2mm}

\section{Benchmark for 3D editing}
\label{sec:benchmark}

Although 3D editing has become increasingly popular in recent years, evaluation of different editing techniques has mainly depended on subjective judgments and non-standard comparison \eg self-captured, unpublished editing scenes. To the best of our knowledge, there is no standard benchmark for evaluating text-based general-purpose 3D editing. In addition, editing prompts in Instruct-NeRF2NeRF only focus on appearance editing and ignore geometric editing. Thus, we collect and provide a simple benchmark, dubbed \benchmark, for this task based on published synthetic data.

First, we select 35 objects from the Objaverse dataset \cite{deitke2023objaverse} from various categories including: furniture, characters, architecture, weapon, animal, vehicle, etc and combine them with NeRF-synthetic dataset as the input for editing. To generate the target prompt, we classify the editing tasks into three types: 1) editing the appearance/texture without changing the geometry structure of the input, 2) adding or removing an object/part from the input, and 3) translating the shape of a single object/part. For each 3D asset, we use GPT-4 \cite{achiam2023gpt} to generate at least 1 target prompt \footnote{we also include the instruction for each entry for instruction-based editing task} for each type of editing, which results in \textbf{112} editing scenarios in total. While $35$ unique objects and $112$ editing prompts seem relatively small, especially when comparing with the image generation/editing benchmark, it is worth mentioning that the benchmark of a closely related task - 3D generation - usually comprises a set of $81$ prompts in Dreamfusion \cite{poole2022dreamfusion} or $10$ prompts in ProlificDreamer \cite{wang2023prolificdreamer}. 
\section{Image Editing Experiments}
\begin{figure}[t]
\centering
\includegraphics[width=\textwidth]{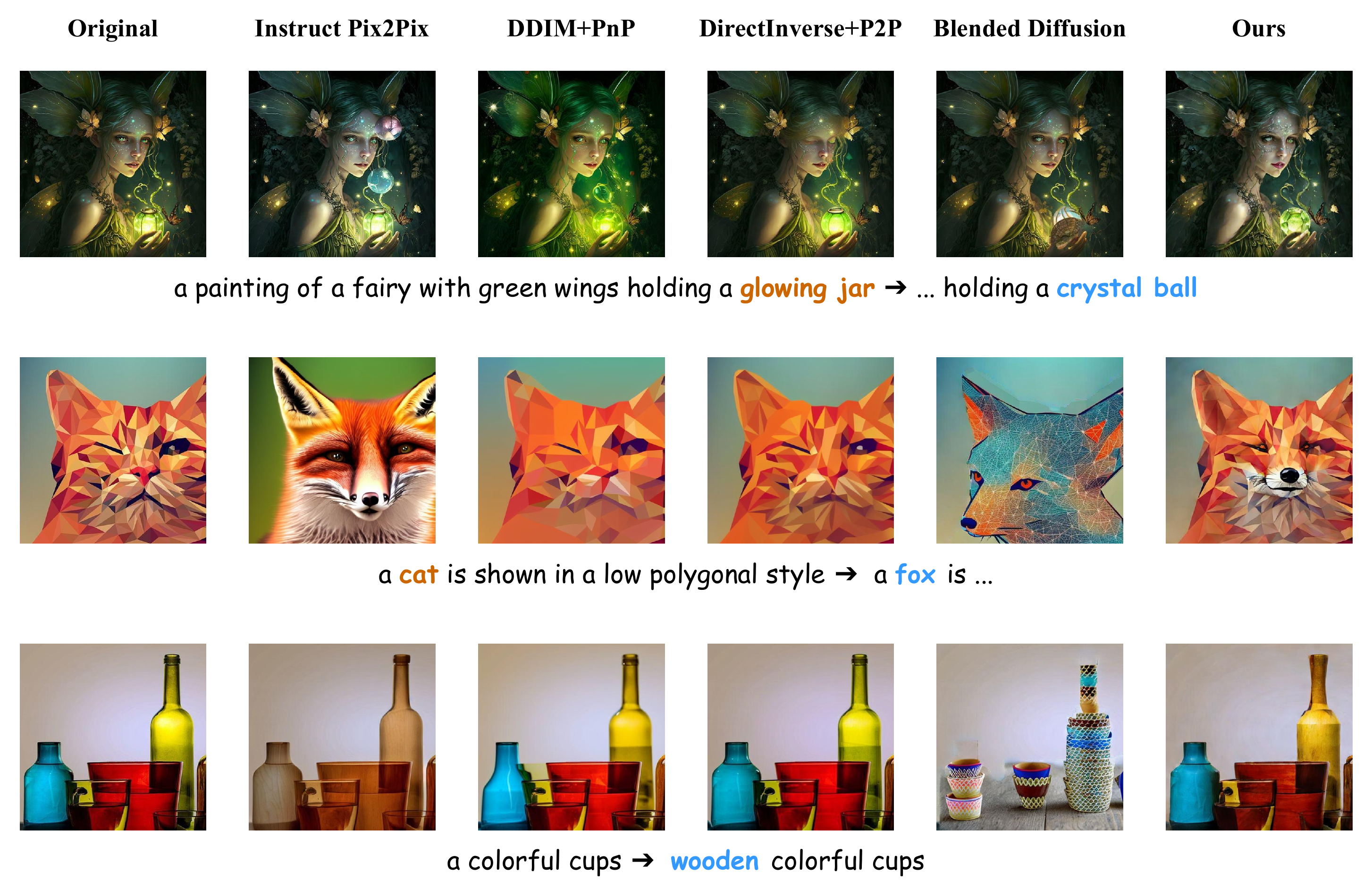}
\vspace{-5mm}
\caption{\textbf{Results of 2D editing with our method and baselines}. We use the results from DirectInversion \cite{ju2023direct} for every method. }
\label{fig:2d_editing_comparison}
\vspace{-3mm}
\end{figure}

\subsection{Experiment Setup}

\paragraph{Dataset and Metrics.} We demonstrate the effectiveness of our method for 2D editing on the PIE-Bench \cite{ju2023direct}. More concretely, PIE-Bench consists of 700 images from diverse scenes, spanning 4 categories (human, animal, indoor, outdoor) and 10 unique editing tasks, including adding objects, deleting objects, changing background and foreground, etc. Similar to \cite{ju2023direct}, we employ 7 different metrics, measuring both the edit fidelity according to the text prompt and the consistency between the source and edited images.

\paragraph{Model Details and Hyperparameters.} We use the \textit{Stable-Diffusion-2-1-base} as the pretrained T2I diffusion model. The secondary diffusion model is simply a LoRA instead of ControlNet as it can effectively model the score of distribution of source and edited images. We use a CFG value of $7.5$ for the pretrained T2I diffusion model and $1.0$ for the LoRA. The value of $\lambda$ is set to $0.4$. Each image is optimized for $1000$ step with the SGD optimizer \cite{kingma2014adam}. We use a cosine learning rate schedule, and the learning rate is set to $0.01$. While we set a fixed number of optimization steps to $1000$ to allow large-spatial structure editing, we observe that most of the editing can be done in $300-500$ steps, which is around 2-4 minutes on a single NVIDIA A100 GPU. 

We compare our method with other editing techniques including: 1) Inversion-based approaches: DDIM-Inversion \cite{song2020score}, Null-text inversion \cite{mokady2023null}, Negative-Prompt Inversion\cite{miyake2023negative}, StyleDiffusion \cite{li2023stylediffusion}, DirectInversion \cite{ju2023direct}, 2) Model-based editing: Instruct Pix2Pix \cite{brooks2023instructpix2pix}, Instruct Diffusion \cite{geng2023instructdiffusion}. 

\paragraph{Experiment Results.}
We report the performance on structure retention, background preservation, and edit clip similarity of all methods in Table~\ref{tab:2d_benchmark}. A quantitative comparison is presented in Figure \ref{fig:2d_editing_comparison}. \emph{We did not tune the learning rate, $\lambda$, learning rate schedule or weight decay}. Regardless, our approach obtains competitive performance with SoTA image editing approaches - consistently achieving the best or second-best performance across all metrics.

\begin{table}[t]
\small
\centering
\renewcommand\arraystretch{0.8}
\setlength{\tabcolsep}{0.3mm}
\renewcommand{\arraystretch}{1.1}
\resizebox{\textwidth}{!}{
\begin{tabular}{c|c|c|cccc|cc}
\toprule
\multicolumn{2}{c|}{\textbf{Method}} & \textbf{Structure} & \multicolumn{4}{c|}{\textbf{Background Preservation}} & \multicolumn{2}{c}{\textbf{CLIP Similariy}} \\ \midrule
\textbf{Inverse} & \textbf{Editing} & \textbf{Distance}$_{^{\times 10^3}}$ $\downarrow$ & \textbf{PSNR} $\uparrow$ & \textbf{LPIPS}$_{^{\times 10^3}}$ $\downarrow$ & \textbf{MSE}$_{^{\times 10^4}}$ $\downarrow$ & \textbf{SSIM}$_{^{\times 10^2}}$ $\uparrow$ & \textbf{Whole} $\uparrow$ & \textbf{Edited} $\uparrow$ \\ \midrule
\multicolumn{2}{c|}{\textbf{Instruct Pix2Pix}} & 57.91 & 20.82 & 158.63 & 227.78 & 76.26 & 23.61 & 21.64 \\
\multicolumn{2}{c|}{\textbf{Instruct Diffusion}} & 75.44 & 20.28 & 155.66 & 349.66 & 75.53 & 23.26 & 21.34 \\ \midrule
\textbf{DDIM} & \textbf{P2P} & 69.43 & 17.87 & 208.80 & 219.88 & 71.14 & 25.01 & \secondbest{22.44} \\
\textbf{NT} & \textbf{P2P} & 13.44 & 27.03 & 60.67 & 35.86 & 84.11 & 24.75 & 21.86 \\
\textbf{NP} & \textbf{P2P} & 16.17 & 26.21 & 69.01 & 39.73 & 83.40 & 24.61 & 21.87 \\
\textbf{StyleD} & \textbf{P2P} & \secondbest{11.65} & 26.05 & 66.10 & 38.63 & 83.42 & 24.78 & 21.72 \\
\textbf{DirectInversion} & \textbf{P2P} & \secondbest{11.65} & \secondbest{27.22} & \best{\textbf{54.55}} & \secondbest{32.86} & \best{\textbf{84.76}} & 25.02 & 22.10 \\
\textbf{DDIM} & \textbf{MasaCtrl} & 28.38 & 22.17 & 106.62 & 86.97 & 79.67 & 23.96 & 21.16 \\
\textbf{DDIM} & \textbf{P2P-Zero} & 61.68 & 20.44 & 172.22 & 144.12 & 74.67 & 22.80 & 20.54 \\
\textbf{DDIM} & \textbf{PnP} & 28.22 & 22.28 & 113.46 & 83.64 & 79.05 & \best{25.41} & \best{22.55} \\ \midrule
\multicolumn{2}{c|}{\textbf{Ours}} & \best{11.02} & \best{27.39} & \secondbest{56.10} & \best{28.72} & \secondbest{84.75} & \secondbest{25.33} & 22.21 \\ \bottomrule
\end{tabular}
}
\vspace{1mm}
\caption{Quantitative comparison of \algo versus 2D image editing baselines. We highlight the \best{\textbf{best}} and \secondbest{second best} results on each metric.}
\label{tab:2d_benchmark}
\vspace{-5mm}
\end{table}

\section{NeRF Editing Experiments}
\label{sec:experiment}
\subsection{Experiment Setup}
We conduct experiments on both real-world datasets from IN2N (IN2N) \cite{haque2023instruct} and synthetic \benchmark benchmark. As discussed in Sec \ref{sec:benchmark}, we focus on 1) editing texture/style without changing structure, 2) adding/removing subtle details, and 3) transforming global structure. \textbf{For simplicity, we only edited the 3D model in a single stage unless otherwise stated}. For detailed hyperparameters and experiment setups, we refer the reader to supplementary documents.

\begin{figure}[t]
\vspace{-1mm}
\includegraphics[width=\textwidth]{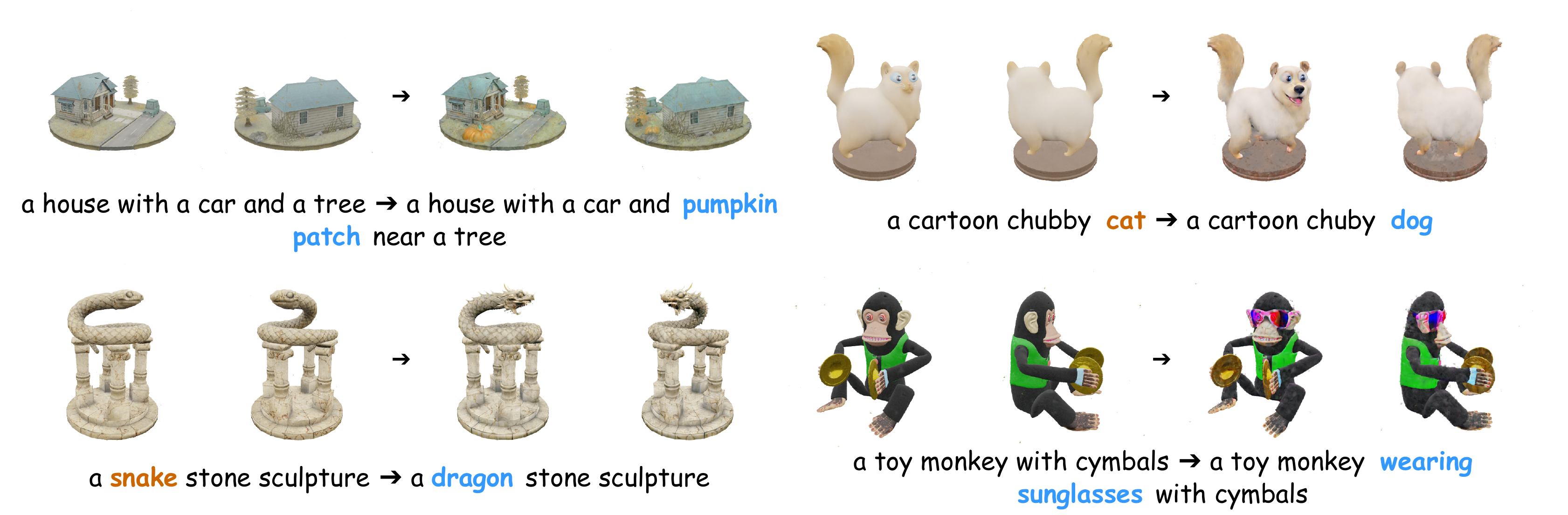}
\vspace{-1mm}
\caption{\algo results on the \benchmark benchmark.}
\vspace{-6mm}
\end{figure}

\begin{figure}[t]
\centering
\includegraphics[width=\textwidth]{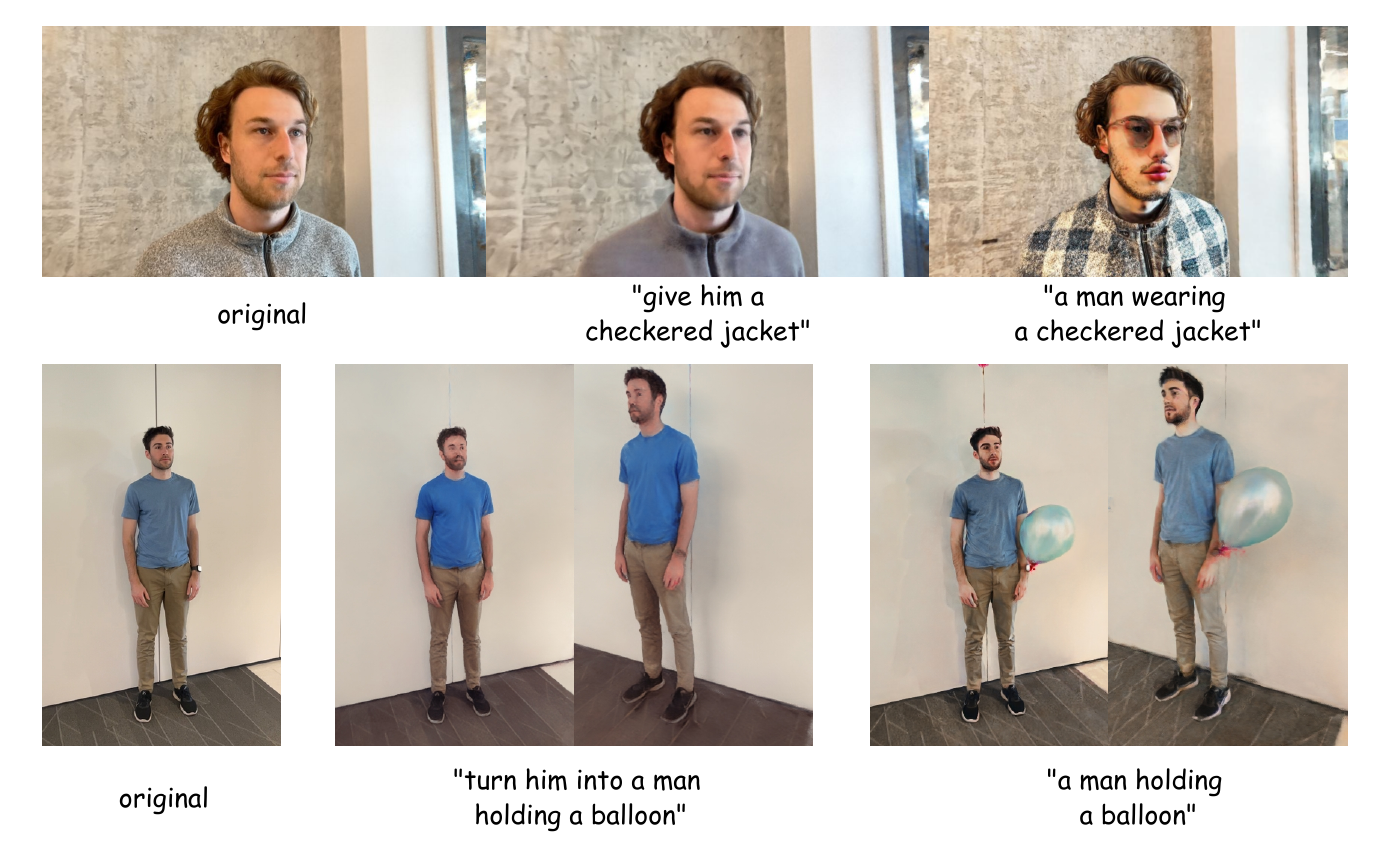}
\vspace{-3mm}
\caption{Real-world NeRF editing over the \textit{face} and \textit{person-small} scenes from the IN2N dataset. The leftmost images are the original images, followed by IN2N and our method \algo.}
\label{fig:in2n_more}
\vspace{-3mm}
\end{figure}

\paragraph{Baselines.} For the \benchmark benchmark, we compare our method with DDS, while we use Instruct NeRF2NeRF (IN2N), DDS and Posterior Distillation Sampling (PDS) \cite{koo2023posterior} as baselines for the IN2N dataset. We re-implemented the DDS loss within the ThreeStudio framework, and for the IN2N and PDS we use the default configuration from their official implementation. In general, we do not extract and fine-tune mesh from NeRF in our comparison.
\vspace{-5mm}

\subsection{Results on \benchmark benchmark.} 
\begin{figure}[t]

\includegraphics[width=\textwidth]{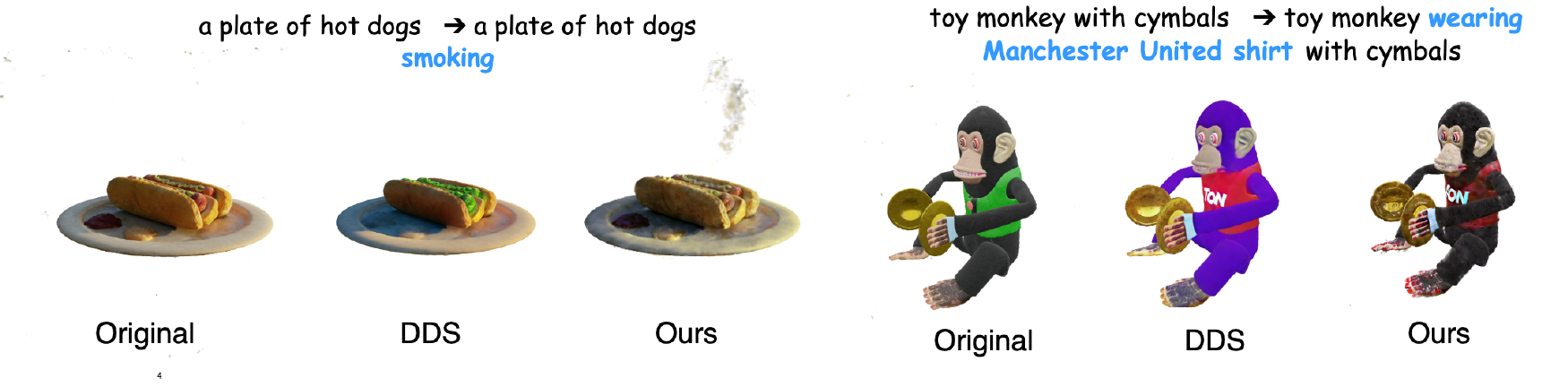}
\caption{Editing on two scenes from the \benchmark benchmark. DDS tends to produce saturated color while \algo always maintains fidelity texture on objects.}
\label{fig:dds_vs_our}
\vspace{-3mm}
\end{figure}

\noindent
\begin{minipage}{0.55\textwidth}
We show the quantitative comparison with the DDS approach on the \benchmark benchmark in Table \ref{tab:dds_vs_our}. We apply the same setup for both DDS and \algo as described in the above section. We render 120 views from each editing task and calculate the CLIP \cite{radford2021learning} and LPIPS score \cite{zhang2018unreasonable} for each 3D model. 
\end{minipage}%
\hfill
\begin{minipage}{0.4\textwidth}
\footnotesize
    \centering
    \begin{tabular}{lcc}
    \hline
    & \textbf{CLIP} ($\uparrow$) & \textbf{LPIPS} ($\downarrow$) \\ \hline
    \textbf{DDS} \cite{hertz2023delta} & 0.242         & 0.1918     \\ 
    \textbf{Ours} & 0.245     & 0.1133     \\ \hline
    \end{tabular}
    \captionof{table}{Quantitative evaluation on the \benchmark benchmark.}
    \label{tab:dds_vs_our}
\end{minipage}

The final score is averaged across all 3D models. Our method achieves better text-image alignment and is also better at preserving original features. In contrast, DDS tends to yield saturated color, as can be observed from the qualitative results in Figure \ref{fig:dds_vs_our}. Additional results can be found in the appendix.

\subsection{Results on IN2N dataset.}
\noindent
\begin{minipage}{0.4\textwidth}
We show the qualitative comparison between \algo and IN2N in Figure \ref{fig:in2n_fangzhou} and Figure \ref{fig:in2n_more}. Compared to IN2N, our method is better at preserving the source details, such as the background and the color and texture of the shirt. Moreover, in the cases where IN2N fails, such as geometric editing and delicate texture as in Figure \ref{fig:in2n_more}, our method still can produce high-quality and consistent editing.
\end{minipage}
\hfill
\begin{minipage}{0.55\textwidth} \footnotesize
    \centering
    \begin{tabular}{lccc} 
    \hline
    & \textbf{CLIP} ($\uparrow$) & \textbf{PSNR} ($\uparrow$) & \textbf{LPIPS} ($\downarrow$) \\
    \hline
    \textbf{IN2N} \cite{haque2023instruct} & 0.238 & 15.84 & 0.356 \\
    \textbf{DDS} \cite{hertz2023delta} & 0.225 & 16.05 & 0.394 \\
    \textbf{PDS} \cite{koo2023posterior} & 0.248 & 17.56 & 0.392 \\
    \textbf{Ours} & 0.250 & 18.60 & 0.312 \\
    \hline
    \end{tabular}
    \captionof{table}{Quantitative evaluation on IN2N data. We follow IN2N and use 9 editing settings of \textit{face} and 3 editing settings of \textit{bear} for evaluation. IN2N and PDS are reproduced from the official implementation with default hyper-parameters.}
    \label{table:in2n_quant}
\end{minipage}%

To further emphasize our ability to generate editing that is consistent with the original image, we conduct the quantitative comparison between \algo and IN2N. We follow the evaluation protocol of Intruct-NeRF2NeRF and select 12 edits from two different scenes named \textit{face} and \textit{bear}. We then perform editing with the two methods and report the CLIP Score, PSNR, and LPIPS. The results, presented in Table \ref{table:in2n_quant} show that \algo matches IN2N in CLIP Score but has a significantly higher PSNR and lower LPIPS. This demonstrates \algo's proficiency in manipulating the NeRF model according to text prompts while preserving the unrelated details intact.

\subsection{Ablation study on the choice of $\lambda$}

\begin{figure}[t]
\centering
\includegraphics[width=\textwidth]{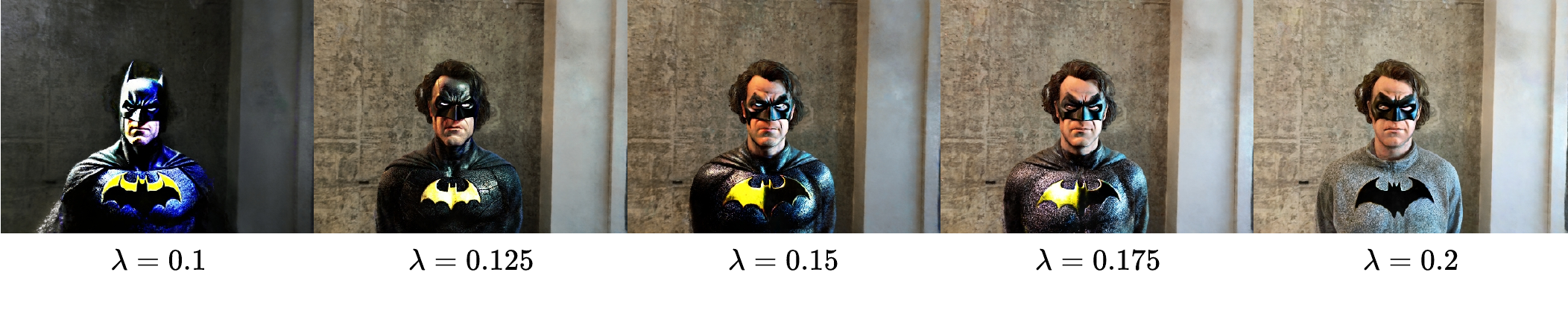}
\vspace{-5mm}
\caption{The edited NeRF varies as $\lambda$ changes. The target prompt is "\emph{Batman}".}
\label{fig:ab_lambda}
\vspace{-0.2cm}
\end{figure}

The parameter $\lambda$ is used by \algo to balance the similarity to the original structure and the alignment with the desired text prompt. By adjusting the value of $\lambda$ in the 3D editing task, as shown in Figure \ref{fig:ab_lambda}, we can explore how it influences this balance. When $\lambda$ is set low, the original image's details, such as the background, clothing, and the appearance of the person, are almost entirely lost. However, as $\lambda$ increases, more features of the original image are preserved, but the strength of the edits decreases. This empirical evidence aligns well with our formulation in Section \ref{sec:proposed_method}.

\section{Related Work}

\paragraph{Text-to-3D Generation} Recent advancements in score-based generative models, particularly diffusion models \cite{song2020score, ho2020denoising}, have significantly impacted text-to-image synthesis. Numerous recent research aims to leverage and transfer the prior of 2D generative model to other modalities such as video and 3D generation.  In the realm of 3D generation, the pioneering work DreamFusion \cite{poole2022dreamfusion} introduced the Score Distillation Sampling (SDS) loss to utilize 2D priors for generating 3D NeRF from text prompts. This loss was also proposed in the concurrent work Score Jacobian Chaining \cite{wang2023score}, and has become extensively popular for generating 3D models \cite{lin2023magic3d, liu2023zero, chen2023fantasia3d, yi2023gaussiandreamer}. Nevertheless, a significant drawback of the SDS loss is its reliance on a high classifier guidance scale, leading to issues such as color over-saturation and excessive smoothing. ProfilicDreamer \cite{wang2023prolificdreamer} proposed Variational Score Distillation (VSD) to solve this problem. VSD optimizes the KL divergence between the distribution of the images rendered from the 3D model and the pretrained 2D diffusion model and produces 3D models with higher diversity and fidelity compared to SDS.

\paragraph{Neural Field Editing} NeRF editing poses significant challenges as it requires manipulating both the geometry and appearance of a 3D scene. EditNeRF \cite{liu2021editing} introduces a method to address this problem by updating the latent codes of a conditional radiance field. Other work \cite{chiang2022stylizing, huang2022stylizednerf, nguyen2022snerf} study the problem of 3D stylization based on a reference style. More recent research \cite{wang2022clip, wang2023nerf, gordon2023blended, gao2023textdeformer} employ visual language representation of CLIP \cite{radford2021learning} to steer NeRF editing using text or image prompts, enhancing the intuitiveness of the editing process. Instruct-NeRF2NeRF \cite{haque2023instruct} is one of the first methods to leverage 2D diffusion prior \cite{brooks2023instructpix2pix} for text-based NeRF editing with an iterative dataset update scheme. Building on its foundation, subsequent research further improves this approach by specifying a mask of the editing region \cite{mirzaei2023watch} or incorporating a view consistent between 2D edited images \cite{song2023efficient, dong2024vica}.  As an alternative to the dataset update approach, other studies \cite{li2023focaldreamer, zhuang2023dreameditor, kim2024collaborative, koo2023posterior} have adopted the SDS loss and its variants to directly leverage a 2D diffusion model for manipulating a NeRF scene. Posterior Distillation Sampling (PDS) \cite{koo2023posterior} is the closest to our work, in which they also solve the problem of identity preservation for DDS. However, their method is based on matching the stochastic latent of the source and target, while we propose an explicit score distillation term as a regularizer for DDS.

\vspace{-0.25cm}
\section{Conclusion}

In this work, we introduce \algo, a novel method for general-purpose image and NeRF editing by leveraging the prior from pre-trained diffusion models. Our method successfully tackles the problem of content shifting observed in the DDS loss and keeps important aspects of the input unchanged. Furthermore, we introduce the synthetic \benchmark benchmark, which includes a variety of objects and prompts for editing.  Through extensive experiments on both tasks of image and NeRF editing, we demonstrate that our model can create high-quality edits while keeping the crucial information from the original data intact.

\paragraph{Limitation and Future Work.} A primary drawback of \algo, when compared to DDS, lies in the necessity to train two additional LoRA networks. This requirement leads to slower training times and a higher demand for computational resources for \algo. We hope to resolve this challenge in future work. Another promising direction would be applying \algo to Gaussian Splatting \cite{kerbl20233d}.
\bibliographystyle{splncs04}
\bibliography{egbib}
\newpage
\appendix

\section{More Implementation Details}

\subsection{NeRF editing.}

\paragraph{\benchmark benchmark.}
For each object in the \benchmark benchmark, we first rescale them to the bounding box with the radius of $2$, which is approximately the same as NeRF synthetic. Our codebase is built on top of the open-sourced ThreeStudio \cite{threestudio2023}. First, we render 100 views uniformly on the upper hemisphere of the input mesh to train an Instant-NGP \cite{mueller2022instant} model of each scene. Specifically, we use Adam optimizer and train it for $10000$ steps. After that, we edit the images by optimizing the NeRF models for $15000$ with the Adam optimizer and a learning rate of $0.01$ for the hash grid encoder and $0.0001$ for the LoRA layers. For the hyperparameter $\lambda$, we use $\lambda = 0.4$ across our experiments on the \benchmark benchmark. We use high-resolution training and an annealed distilling time schedule to improve the visual quality similar to \cite{wang2023prolificdreamer}. 
It is worth mentioning that we do not use view-dependent prompting similar to prior work \cite{poole2022dreamfusion, wang2023prolificdreamer} as the \benchmark benchmark does not align every object to a canonical coordinate system.

\paragraph{IN2N dataset.} We conduct experiments on NeRFStudio \cite{tancik2023nerfstudio}, which is optimized for real-world unbounded scenes. Specifically, for each scene, we train the \texttt{nerfacto} model from NeRFStudio for $10000$ steps. We then train the model with Adam optimizer with the learning rate of $0.001$ for the NeRF model and $0.0001$ for the LoRA layers. We fix $\lambda=0.2$ across our experiments.

As stated in the main paper, we use ThreeStudio for synthetic object editing. We use a batch size of $1$ due to the computational limit. A large batch size of NeRF may improve the generated quality and convergence speed, we leave it in future work. We use high-resolution \ie $512\times 512$, for all $10000$ editing steps. The bounding box of object-centric data \ie \benchmark benchmark is roughly $2.0$. At each training step, we random a camera poses with elevation in the range of $[10; 60]$, field-of-view (FoV) in the range of $[40; 70]$, camera distance in the range of $[3;4]$. For evaluation, we use a fixed camera distance of $4$, FoV of $50$, elevation of $15$ and render $120$ images around the object while varying the azimuth. We calculate the CLIP and LPIPS scores of each method for all rendered images. In addition, we set the weighting of loss function $\omega(t) = \sigma_t^2$.

For $\mathbf{D}_\phi$ and $\mathbf{D}_\psi$, we parameterize both of them Depth-conditioned ControlNet. We use the Classifier-Free Guidance (CFG) value of $1$ for source and target diffusion models.

\section{Benchmark constructing}
We leverage the in-context learning capability of GPT-4 to generate target and source prompts for each object. More concretely, the context used for prompting the generation of GPT-4 is illustrated in Table \ref{tab:context}. For each 3D object, we first render a front-view image of the mesh and use the dialogue as in Table \ref{tab:generate_json_entry} to produce \textbf{three} types of editing including texture editing, adding/removing part (local editing) and shape translation (global editing).
\begin{table}
  \begin{minipage}{0.99\textwidth}
\centering  

\begin{tabular}{l p{10.5cm} }
\toprule
 \multicolumn{2}{l}{\bf Editing task generation example}  \\
\midrule
&  \includegraphics[height=4.5cm]{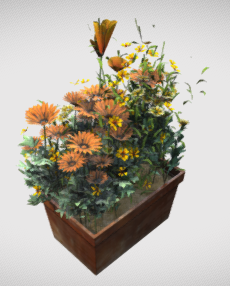} \\

User & path: objaverse\_synthetic/e114ab027906482abe7daa04eef60e95 \\

GPT-4 &  
\begin{lstlisting}[breaklines=true, basicstyle=\ttfamily\footnotesize]
{
    "path": "objaverse_synthetic/e114ab027906482abe7daa04eef60e95",
    "source_prompt": "Flower box with blooms",
    "target_prompt": "Flower box with autumn leaves",
    "type": "texturing"
},
{
    "path": "objaverse_synthetic/e114ab027906482abe7daa04eef60e95",
    "source_prompt": "Flower box with blooms",
    "target_prompt": "Flower box with a butterfly",
    "type": "addition"
},
{
    "path": "objaverse_synthetic/e114ab027906482abe7daa04eef60e95",
    "source_prompt": "Flower box with blooms",
    "target_prompt": "Herb garden box",
    "type": "translation"
}

\end{lstlisting}

\end{tabular}
\captionof{table}{Illustrate the dialogue between user and GPT-4 to generate three editing entries for a given 3D model. We render a front-view image of an Objaverse object. Then, we send that image along with the path to the image (which also contains its unique ID) to GPT-4. The model will return three JSON entries corresponding to different types of editing for the input object. }
\vspace{-5mm}
\label{tab:generate_json_entry}  
  \end{minipage}
\end{table}

\begin{table*}[t!]\centering
\begin{minipage}{1.0\columnwidth}\vspace{0mm}    \centering
\begin{tcolorbox} 
    \centering
   
      \footnotesize
    \begin{tabular}{p{0.97\columnwidth} c}
   \VarSty{ {\bf Context} } &\\
You are an assistant helping me create a dataset for image editing. You will be given an image. First, you will return a short description of that image. Then, you have to specify a target prompt for editing the given image. The target prompt should be mostly similar to the source prompt. The target must change the source image in one of the following ways: \\

1. changing appearance: changing only the texture or the style of the given image without varying its structure. For example, a red shirt to a blue shirt, a watercolor dragon to a photorealistic dragon, a horse to a zebra, a man to a clown, etc \\

2. adding a new object to the scene: almost keeping the given image intact (but you can delete part of it) and adding a new object to the scene. For instance, a duck to a duck wearing sunglasses, a dragon to a dragon wearing a party hat, a flamingo to a flamingo on a skateboard, etc\\

3. translate the shape: you can change the structure of the image, but it should change to something similar in terms of geometry structure: a cat to a dog, a knight riding a horse to a knight riding an elephant, etc

Make sure to diversify the target prompt beyond my example. The target prompt should be concise and only have 1 change for each. \\

The input will be an image and a path to that image, you are expected to return a JSON text similar to this:
\begin{lstlisting}[breaklines=true, basicstyle=\ttfamily\footnotesize]
{
    "path": "objaverse_synthetic/b363f8f4f7394ddb9d5b3a337f2f7fc7",
    "source_prompt": "A pink flamingo standing on one leg",
    "target_prompt": "A pink flamingo with rainbow feathers standing on one leg",
    "type": "texturing"
},
{
    "path": "objaverse_synthetic/b363f8f4f7394ddb9d5b3a337f2f7fc7",
    "source_prompt": "A pink flamingo standing on one leg",
    "target_prompt": "Photorealistic pink flamingo standing, wearing a gold necklace",
    "type": "addition"
},
{
    "path": "objaverse_synthetic/b363f8f4f7394ddb9d5b3a337f2f7fc7",
    "source_prompt": "A pink flamingo standing on one leg",
    "target_prompt": "A pink crane standing on one leg",
    "type": "translation"
},
    
\end{lstlisting}

    \end{tabular}
\end{tcolorbox}
\caption{Illustrate the context for the dialogue between the user and ChatGPT 4 to generate the editing benchmark.}
    \label{tab:context}
\end{minipage}
\end{table*}

\section{Algorithm}
Algorithm \ref{algo} outlines our pipeline for 3D editing with \algo.

\begin{algorithm}[H]
\label{algo}
    \caption{NeRF editing with \algo}
    \SetKwInOut{Input}{Input}
    \SetKwInOut{Output}{Output}
    \SetKwFunction{to}{to}
    \Input{Original NeRF $g(\theta)$, pretrained diffusion model $\mathbf{D}_\kappa$, text prompts $\mathbf{c}_{tgt}$ and $\mathbf{c}_{src}$, $\lambda$.}
    Initialize LoRA networks $\mathbf{D}_\psi$ and $\mathbf{D}_\phi$ from $\epsilon_\kappa$\;
    \While{\text{not reached max\_iterations}}{
        Render a random view  $\mathbf{x}$ from $g_\theta$\;
        Sample $\epsilon \sim \mathcal{N}(0, \mathbf{I})$ and $t \sim U[0, T]$\;
        Add noise $\mathbf{x}_T = \alpha_t \mathbf{x} + \sigma_t \epsilon$\;
        Calculate noise predictions: $\epsilon_\kappa\left(\mathbf{x}_t, t, \mathbf{c}_{tgt}\right)$, $\epsilon_\kappa\left(\mathbf{x}_t, t, \mathbf{c}_{s r c}\right)$,
        $\epsilon_\psi\left(\mathbf{x}_t, t, \mathbf{c}_{s r c}\right)$,
        $\epsilon_\phi\left(\mathbf{x}_t, t, \mathbf{c}_{t g t}\right)$ \;
        Calculate $\nabla_\theta \mathcal{L}(\theta)$ using Eq. \ref{eq:dds_7} and update $\theta$\;
        Update $\mathbf{D}_\psi$ and $\mathbf{D}_\phi$ using diffusion losses from Eq. \ref{eq:ddpm_loss} and Eq. \ref{eq:ddpm_loss_2}\;
    }
    \Return{Edited NeRF model $g(\theta)$.}
\end{algorithm}

\section{Additional results}
\subsection{Effect of ControlNet for $\mathbf{D}_\phi$ and $\mathbf{D}_\psi$}

\begin{figure}[t]
\includegraphics[width=\textwidth]{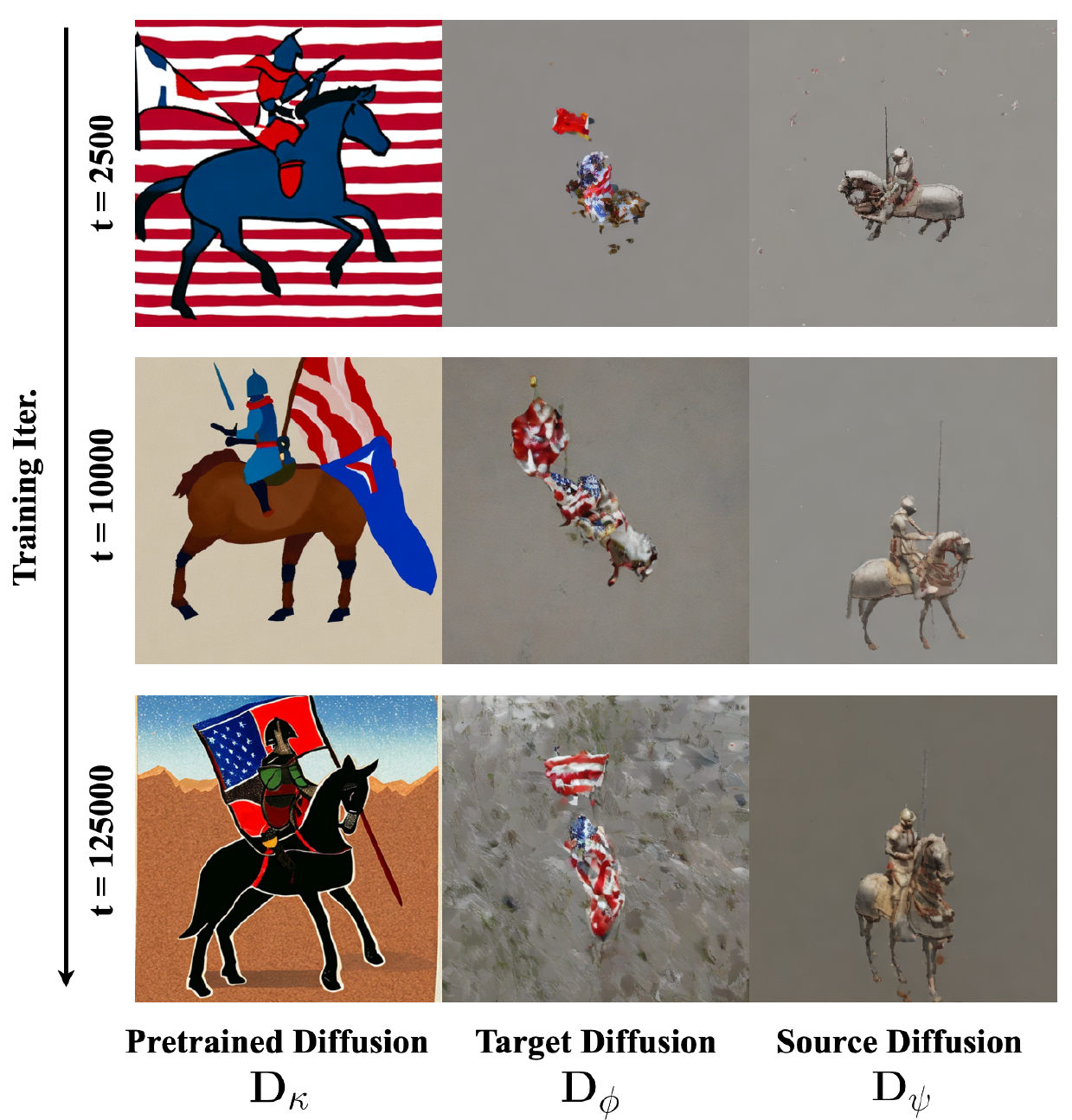}
\caption{Illustrating samples obtained with $50$ steps of DDIM \textbf{without ControlNet} from pre-trained diffusion, source diffusion, and target diffusion at various training steps. The target and source diffusion models are parameterized with LoRA. Since we do not condition the samples from the source and target diffusion on view-pose, they could be from distinct camera poses when sampling. The source prompt is \emph{``A knight riding a horse''}, and the target prompt is \emph{``A knight riding a horse and holding a flag''}. Note that we sample images from the pretrained diffusion model with the target prompt.} 
\label{fig:without_controlnet}
\end{figure}

\begin{figure}[t]
\includegraphics[width=\textwidth]{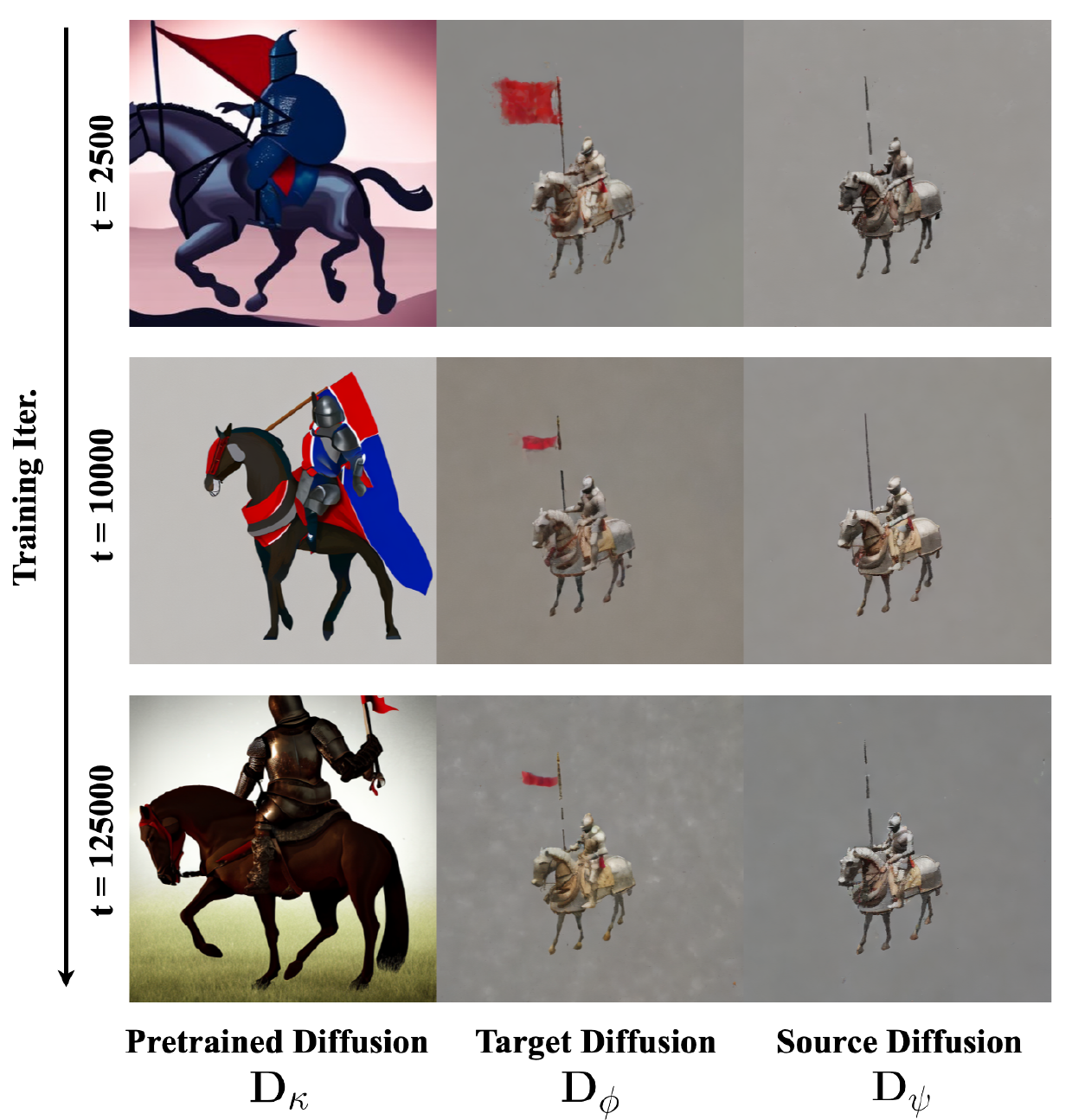}
\caption{Illustrating samples obtained with $50$ steps of DDIM \textbf{with ControlNet} from pre-trained diffusion, source diffusion, and target diffusion at various training steps. The target and source diffusion models are parameterized with depth-conditioned ControlNet. By conditioning the model on depth, samples obtained from source and target diffusion are noticeably better than LoRA. In addition, samples are more consistent than LoRA in terms of view poses. The source prompt is \emph{``A knight riding a horse''}, and the target prompt is \emph{``A knight riding a horse and holding a flag''}. Note that we sample images from the pretrained diffusion model with the target prompt.} 
\label{fig:with_controlnet}
\end{figure}

\begin{figure}[t]
\includegraphics[width=\textwidth]{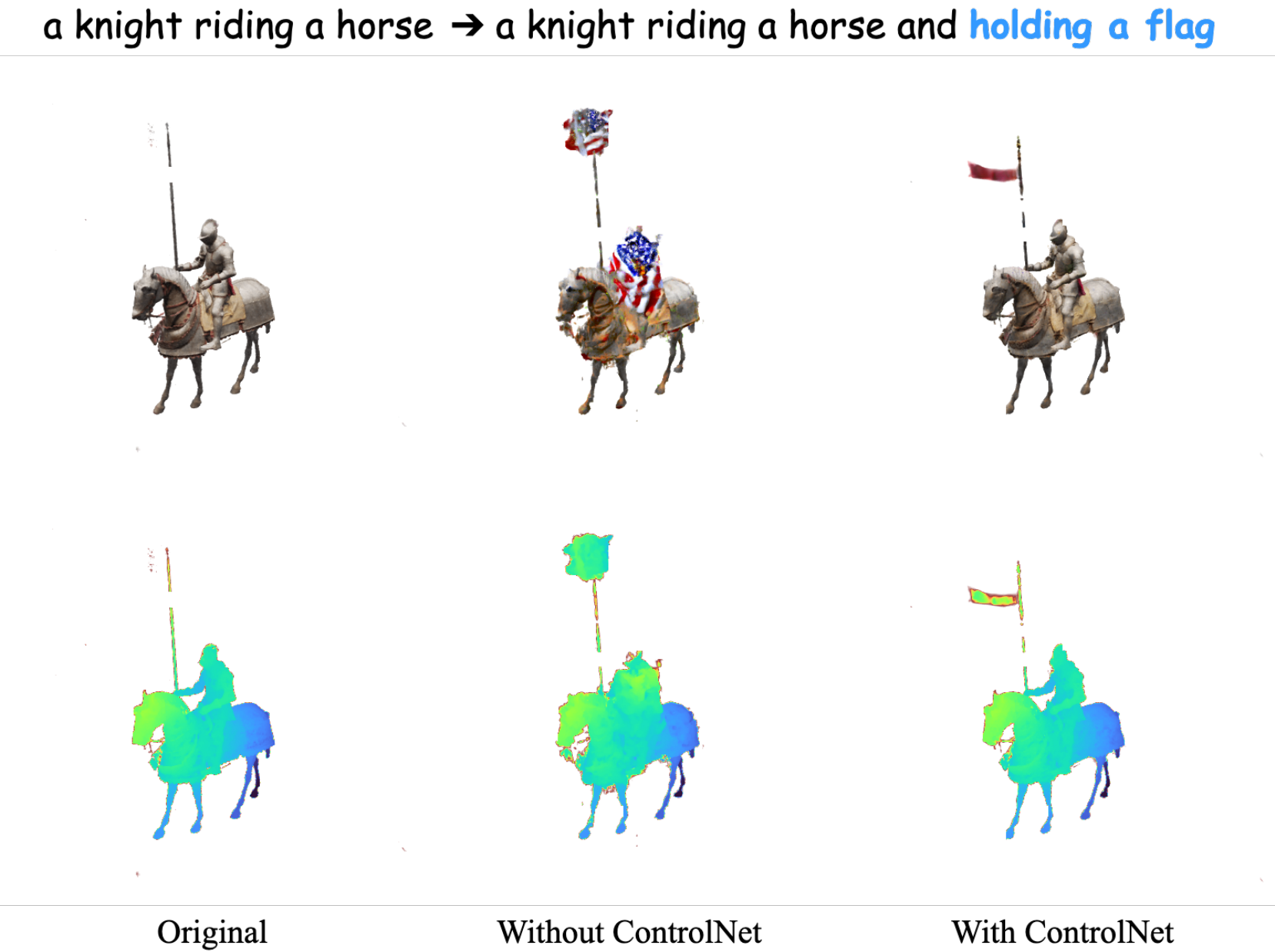}
\caption{Illustrating edited NeRF models with and without ControlNet. The source prompt is \emph{``A knight riding a horse''}, and the target prompt is \emph{``A knight riding a horse and holding a flag''}. ControlNet leads to better consistency with original NeRF and better-edited results.} 
\label{fig:with_and_without_controlnet}
\end{figure}

Although parameterize the source and target diffusion \ie $\mathbf{D}_\psi$ and $\mathbf{D}_\phi$ with LoRA is sufficient in 2D cases, we observe that naively applying that to the 3D case is insufficient. Particularly, without conditioning on view-pose, the source diffusion model can denoise the noisy input from images from arbitrary view-pose. Thus, the reconstruction signal will be rather poor and can lead to Janus/multi-faced problem. To illustrate this phenomenon, we visualize samples from pretrained diffusion $\mathbf{D}_\kappa$, source diffusion $\mathbf{D}_\psi$, and target diffusion $\mathbf{D}_\phi$ with ancestral sampling in Figure \ref{fig:without_controlnet}. We can observe that the view pose varies significantly across sampling. In contrast, ControlNet with depth condition as in Figure \ref{fig:with_controlnet} provides additional cues for denoising step, thus leading to more consistent results (in terms of view poses). Moreover, ControlNet leads to more plausible NeRFs as in Figure \ref{fig:with_and_without_controlnet}.

\subsection{More comparison with DDS}
\begin{figure}[t]
\centering
\includegraphics[width=\textwidth]{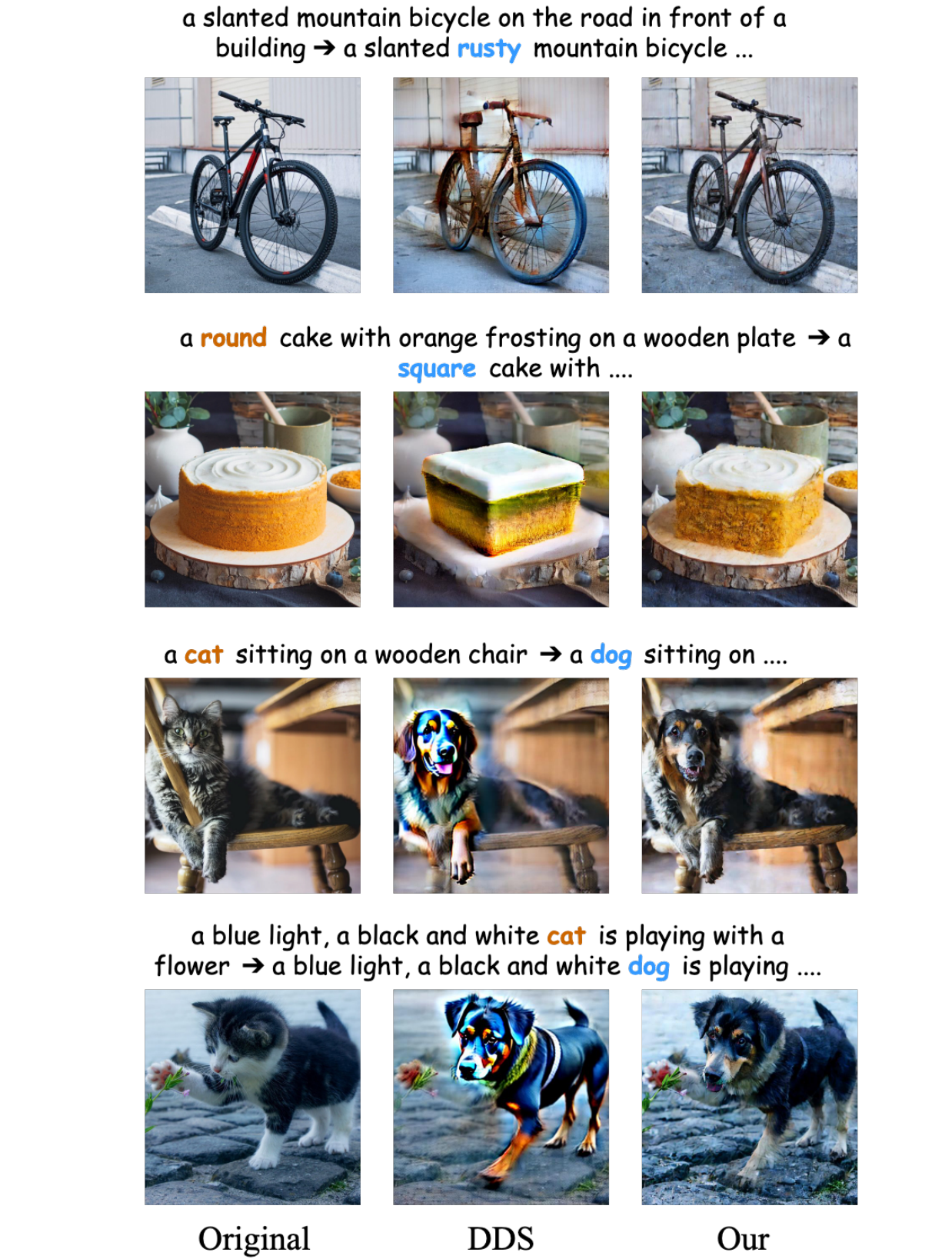}
\caption{Qualitative comparison between DDS and our approach for \textbf{image} editing on PIE-Benchmark. With only optimizing images for $200$, DDS already diverged the edited image from the original. More important, DDS leads to severe over-saturation. In contrast, our approach generates fidelity results while maintaining details from source images.} 
\label{fig:dds_vs_our_2d}
\end{figure}

\begin{figure}[t]
\centering
\includegraphics[width=0.8\textwidth]{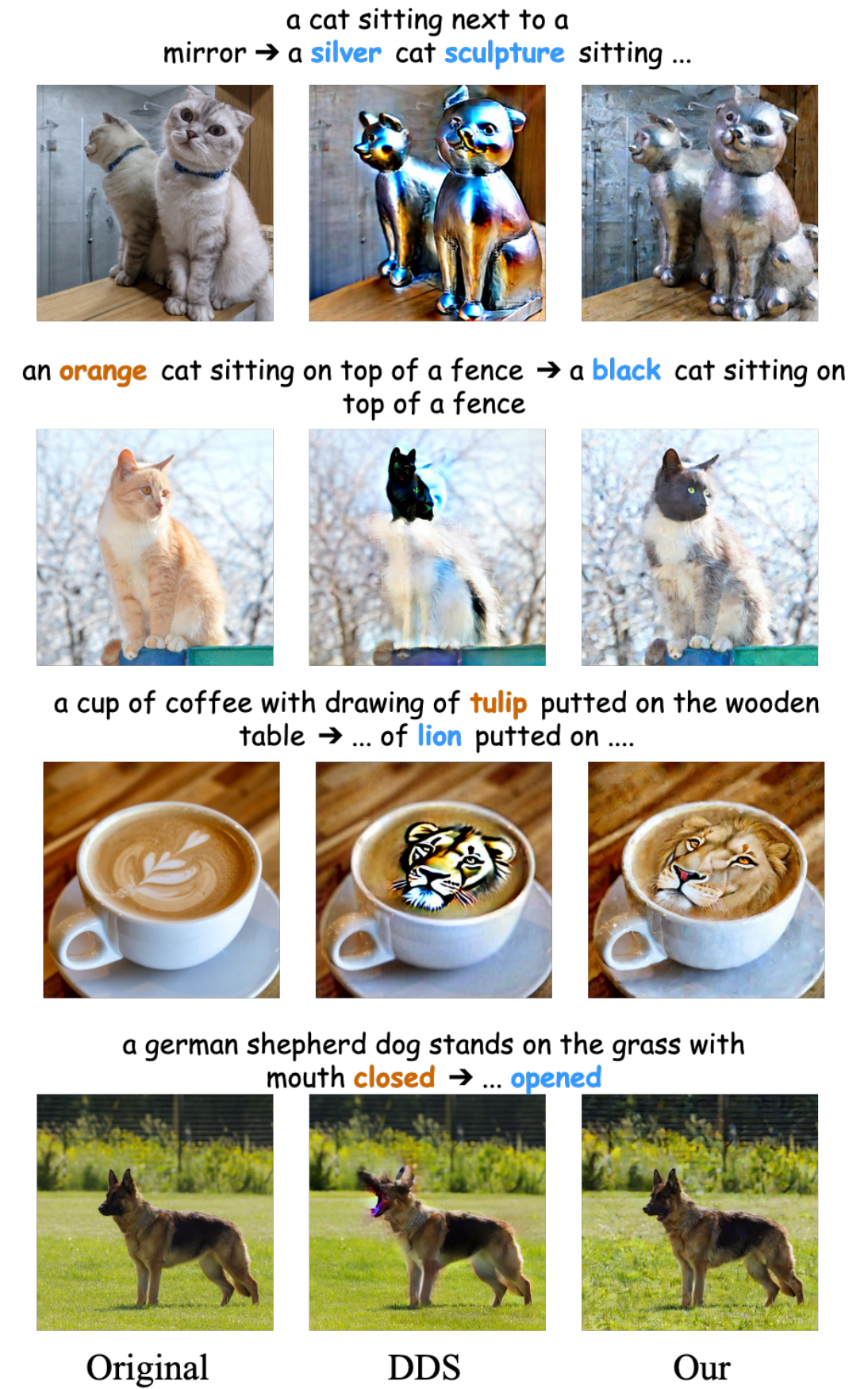}
\caption{Qualitative comparison between DDS and our approach for \textbf{image} editing on PIE-Benchmark.} 
\label{fig:dds_vs_our_2d_p2}
\end{figure}

\begin{figure}[t]
\centering
\includegraphics[width=0.8\textwidth]{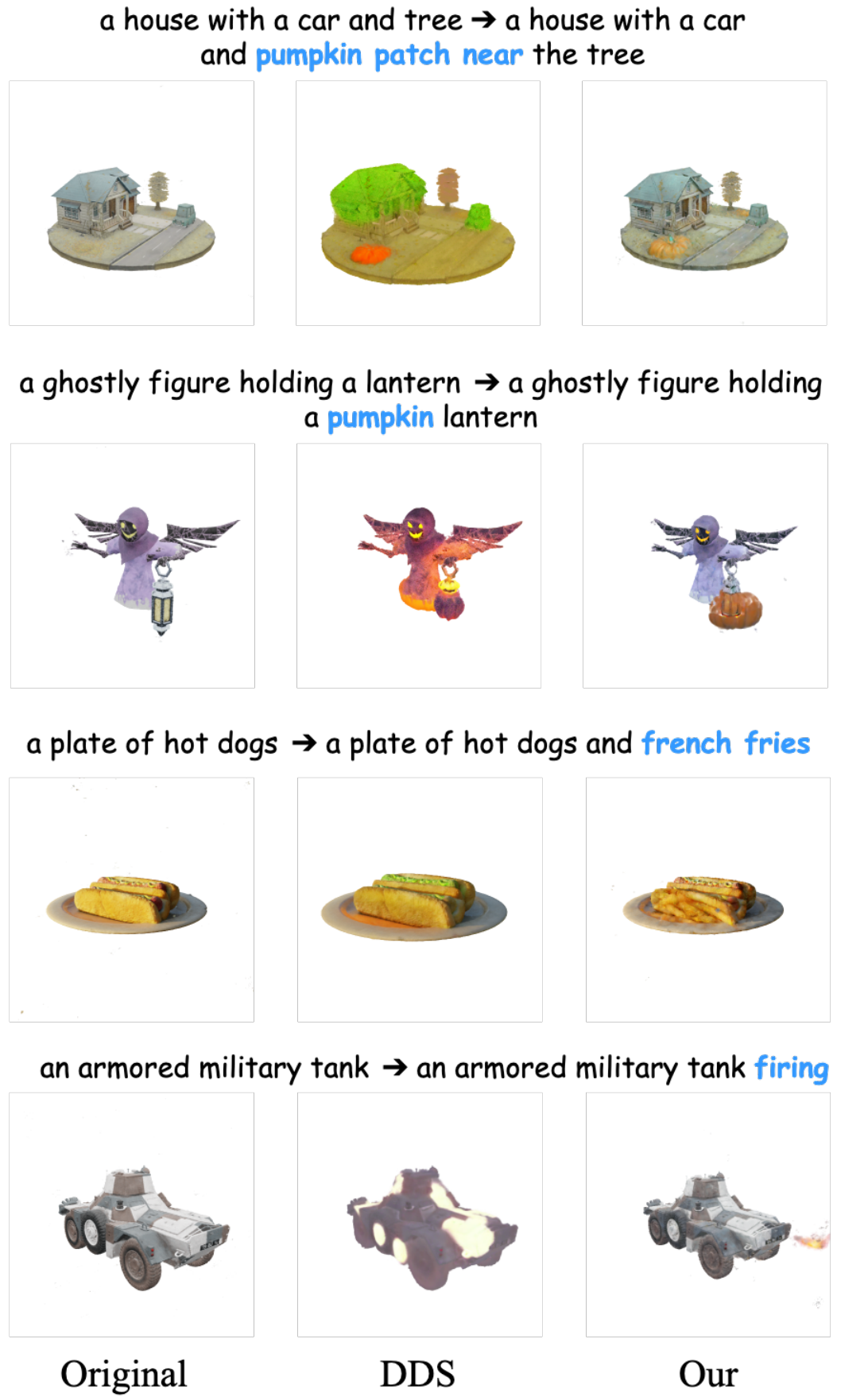}
\caption{Qualitative comparison between DDS and our approach for \textbf{NeRF} editing on \benchmark benchmark. Similar to 2D case, DDS always leads to over-saturation and fails to generate NeRFs aligned with target prompts. In contrast, our approach generates plausible results and does not suffer from over-saturation.} 
\label{fig:dds_vs_our_3d}
\end{figure}

We provide additional qualitative comparisons between DDS and our approach to 2D and 3D. For image editing, we illustrate the edited images from both methods on PIE-Bench in Figure \ref{fig:dds_vs_our_2d} and Figure \ref{fig:dds_vs_our_2d_p2}. Even with only $200$
optimization steps, DDS leads to saturated color and diverges from the original images. On the other hand, our approach is stable with longer training: \algo successfully edits the input image to conform with the target prompt while preserving details from the source image. More important, our approach does not suffer from severe over-saturation. For NeRF editing, we can observe that DDS still encounters the above problems as in Figure \ref{fig:dds_vs_our_3d}. Again, our approach solves these problems and returns satisfactory NeRF models.

\subsection{More qualitative results on IN2N data}
\begin{figure}[t]
\includegraphics[width=\textwidth]{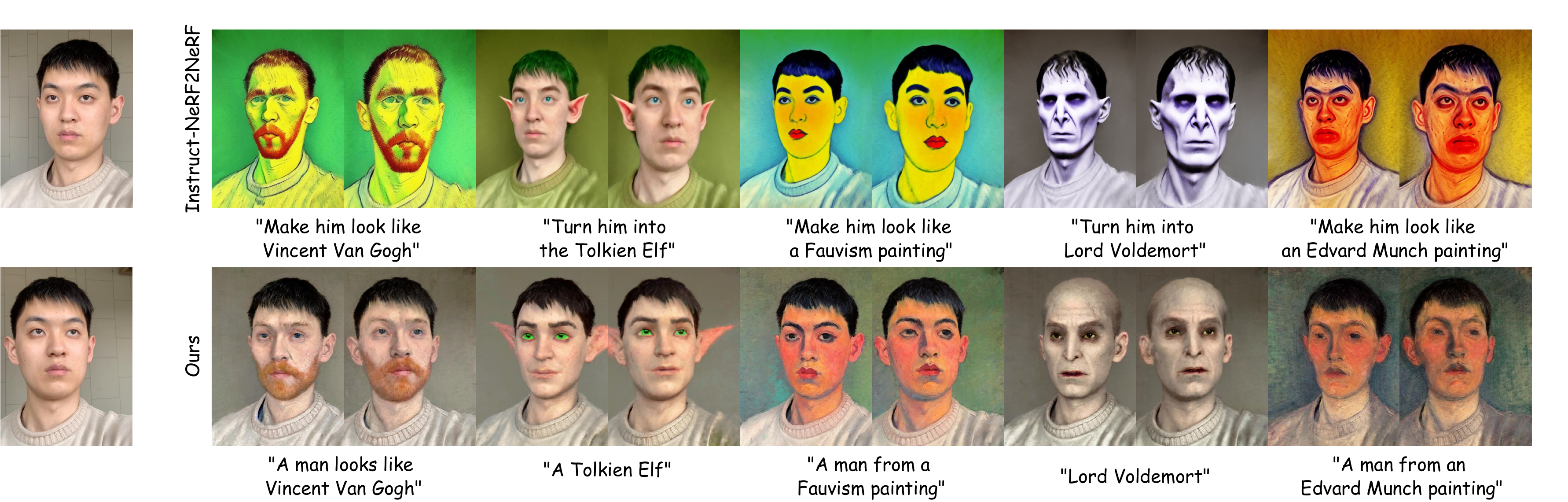}
\caption{Real-world NeRF editing over the \textit{fangzhou-small} scene from the IN2N dataset. The leftmost images are the unedited images, and we show 5 editing results of our method versus the baseline IN2N. Note that the text prompts are different due to the prompt structure of Instruct-Pix2Pix.}
\label{fig:in2n_fangzhou}
\end{figure}

We further show the experimental results on the \textit{fangzhou-small} scene of the IN2N dataset in Figure \ref{fig:in2n_fangzhou}. Compared to the Instruct-NeRF2NeRF method, \algo is able to maintain the structure and identity of the per

\section{Broader Impact}
Our work studies the problem of 2D and 3D editing using a pretrained text-to-image diffusion model. Applications of our method may advance digital content generation, especially in 3D modeling. This may help reduce labor costs in several sectors, including gaming, film production, and AR/VR.

\section{Compute Resources}
All experiments in this paper can be run on a single A6000 GPU with 49GB of memory.

\end{document}